%% file: arxiv_version.tex
\documentclass{article}

\usepackage[preprint]{neurips_2025}

\usepackage[utf8]{inputenc} 
\usepackage[T1]{fontenc}    
\usepackage{hyperref}       
\usepackage{url}            
\usepackage{booktabs}       
\usepackage{amsfonts}       
\usepackage{nicefrac}       
\usepackage{microtype}      
\usepackage{xcolor}         
\usepackage{graphicx}
\usepackage{multirow}
\usepackage{amsmath}
\usepackage{colortbl}
\usepackage{makecell}
\usepackage{pifont}
\usepackage{xspace}
\usepackage[misc,geometry]{ifsym}
\usepackage[capitalize]{cleveref}
\crefname{section}{Sec.}{Secs.}
\crefname{table}{Tab.}{Tabs.}

\def\ourmodel{OpenSeg-R\xspace}

\usepackage{soul} %
\soulregister{\cite}7 %
\soulregister{\citep}7 %
\soulregister{\citet}7 %
\soulregister{\ref}7 %
\soulregister{\cref}7 %
\soulregister{\pageref}7 %
\soulregister{\bf}7

\DeclareMathOperator*{\argmax}{arg\,max}
\title{OpenSeg-R: Improving Open-Vocabulary Segmentation via Step-by-Step Visual Reasoning}

\author{Zongyan Han$^{1}$, 
	Jiale Cao$^2$, 
	Shuo Chen$^3$, 
	Tong Wang$^{1}$, 
	\\
	\textbf{Jorma Laaksonen$^4$,
	Rao Muhammad Anwer$^{1}$}
	\\ $^1$Mohamed Bin Zayed University of Artificial Intelligence, Abu Dhabi, UAE
	\\ $^2$Tianjin University, Tianjin, China  $^3$Nanjing University, Nanjing, China
		\\ $^4$Aalto University, Espoo, Finland
	\\{\tt\small \{zongyan.han, tong.wang, rao.anwer\}@mbzuai.ac.ae}
}

\begin{document}
\maketitle

\input{sec/0_abstract}    
\input{sec/1_intro}

\input{sec/2_relate_work}
\input{sec/3_method}
\input{sec/4_experiments}
\input{sec/5_conclusion}

{
	\small
	\bibliographystyle{plain}
	\bibliography{main}
}


\appendix
\input{sec/7_supp}



\end{document}

%% file: sec/0_abstract.tex
\begin{abstract}

Open-Vocabulary Segmentation (OVS) has drawn increasing attention for its capacity to generalize segmentation beyond predefined categories.
However, existing methods typically predict segmentation masks with simple forward inference, lacking explicit reasoning and interpretability. 
This makes it challenging for OVS model to distinguish similar categories in open-world settings due to the lack of contextual understanding and discriminative visual cues.
To address this limitation, we propose a step-by-step visual reasoning framework for open-vocabulary segmentation, named \textbf{OpenSeg-R}. The proposed OpenSeg-R leverages Large Multimodal Models (LMMs) to perform hierarchical visual reasoning before segmentation. 
Specifically, we generate both generic and image-specific reasoning for each image, forming structured triplets that explain the visual reason for objects in a coarse-to-fine manner.
Based on these reasoning steps, we can compose detailed description prompts, and feed them to the segmentor to produce more accurate segmentation masks.  
To the best of our knowledge, OpenSeg-R is the first framework to introduce explicit step-by-step visual reasoning into OVS.
Experimental results demonstrate that OpenSeg-R significantly outperforms state-of-the-art methods on open-vocabulary semantic segmentation across five benchmark datasets.
Moreover, it achieves consistent gains across all metrics on open-vocabulary panoptic segmentation.
Qualitative results further highlight the effectiveness of our reasoning-guided framework in improving both segmentation precision and interpretability.
Our code is publicly available at \url{https://github.com/Hanzy1996/OpenSeg-R}.
\end{abstract}

%% file: sec/1_intro.tex
\section{Introduction}
\label{sec:intro}
Image segmentation is a fine-grained visual recognition task that involves classifying every pixel in an image.  
This technique has been widely applied in domains such as autonomous driving~\cite{cordts2016cityscapes,prakash2021multi}, medical imaging~\cite{butoi2023universeg,zhao2025foundation}, and remote scene~\cite{liu2024remoteclip,wang2024skyscript}.  
However, conventional segmentation methods are constrained by a predefined set of training classes, which limits their generalization ability to open-world.  
Open-Vocabulary Segmentation (OVS)~\cite{zhao2017open,SPNet,wu2024towards} addresses this limitation by leveraging vision-language models such as CLIP~\cite{clip} to enable segmentation beyond closed-set categories.

Current OVS approaches can be broadly categorized into two frameworks.  
The first employs a one-stage architecture~\cite{maskclip1,xie2024sed,catseg}, which computes similarity maps between image features and textual embeddings of class names via CLIP.  
These maps are then refined through upsampling or embedding transformations, followed by feature fusion to produce segmentation masks.  
The second category utilizes a two-stage framework~\cite{maskclip,ovseg,xu2022simple,jiao2024collaborative}, initially generating class-agnostic mask proposals and subsequently aligning mask-specific features with text embeddings for semantic classification.  

\begin{figure}
\centering
\includegraphics[width=\textwidth]{{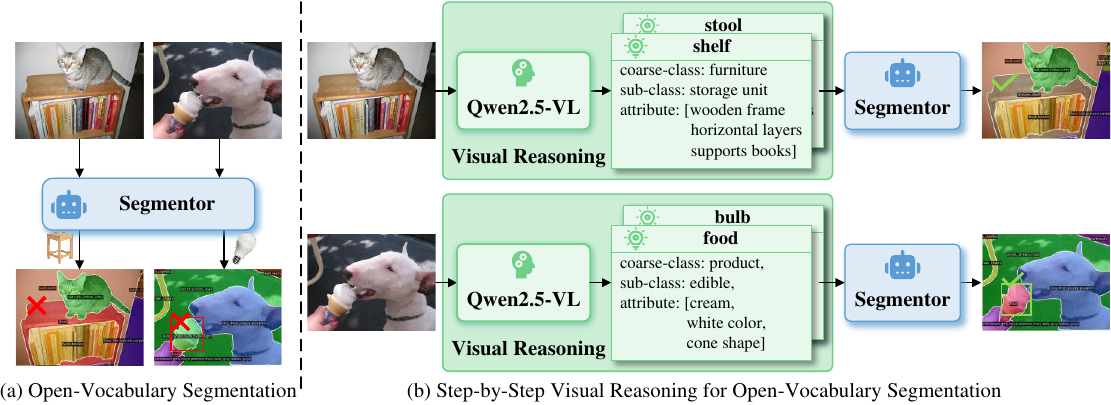}} 
\vspace{-5pt}
\caption{
	Comparison with open-vocabulary segmentor and our method.
(a) Standard open-vocabulary segmentors struggle with visually or semantically similar categories.
(b) Our \ourmodel{} guides the segmentor with hierarchical visual reasons, therefore generating accurate prediction.}
\label{fig:motivation}
\end{figure}

Despite their success, these methods rely solely on associating class name with visual regions for segmentation, without any interpretable explanation or reasoning for segmentation outcomes.
This lack of transparency undermines trust in model decisions and limits adaptability to some complex scenarios and difficult examples. 
For example, as illustrated in ~\cref{fig:motivation}(a), an open-vocabulary segmentor incorrectly classifies a \texttt{shelf} as a \texttt{stool} due to their similar appearance and semantics.

To address this gap, we propose to improve open-vocabulary segmentation via step-by-step visual reasoning, named \textbf{\ourmodel}. It utilizes explicit visual cues and attributes to support segmentation.
These cues offer more detailed and precise knowledge of the objects in the image, thereby enhancing both the accuracy and interpretability of the segmentation process.

Concretely, our framework mimics a human-like reasoning process using a Large Multimodal Model (LMM), such as Qwen2.5-VL~\cite{Qwen2.5-VL}.
The LMM is prompted to generate Image-Specific Reasoning, which begins by producing an overall description of the image and then identifying the presence of relevant object categories.
Based on the observed classes, the LMM progressively explains the association between each object and its corresponding category through coarse-to-fine reasoning steps.
As illustrated in~\cref{fig:motivation}(b), the hierarchical reasoning process first identifies the coarse and sub-category levels, which helps eliminate irrelevant distractor classes (e.g., distinguishing \texttt{food} from \texttt{bulb}).
Subsequently, fine-grained attribute-based descriptions are used to disambiguate visually similar categories (e.g., \texttt{shelf} vs. \texttt{stool}).

To supplement categories that may be overlooked by the LMM, we further introduce a Generic Class Reasoning module.
For each unobserved class, the LMM will generate a coarse-to-fine reason that highlights key features for distinguishing it from visually similar categories.
We then feed both the image-specific and generic reasoning into the segmentor to perform open-vocabulary segmentation.

Our \ourmodel{} not only provides richer and more discriminative semantic context for improved segmentation performance, but also enhances model interpretability in open-world scenarios.
We evaluate \ourmodel{} on open-vocabulary semantic segmentation across five benchmark datasets, and on open-vocabulary panoptic segmentation using a widely adopted benchmark.
On both tasks, our method achieves state-of-the-art performance.

Our contributions can be summarized as below:
\begin{itemize}
    \item We are the first to propose improving open-vocabulary segmentation through step-by-step visual reasoning.
    Our framework, \ourmodel{}, mimics human-like reasoning by leveraging a Large Multimodal Model (LMM).

    \item We introduce two complementary reasoning modules: Image-Specific Reasoning and Generic Class Reasoning, which improve both the accuracy and interpretability of segmentation.
    
    \item Our method achieves state-of-the-art results on five open-vocabulary semantic segmentation benchmarks and the popular open-vocabulary panoptic segmentation benchmark.
    
\end{itemize}

%% file: sec/2_relate_work.tex
\section{Related Work}\label{sec:relate_work}
\subsection{Open-Vocabulary Segmentation}
Open-vocabulary segmentation has emerged recently as its ability to segment unseen categories.
Early efforts in this field focus on aligning visual features with pre-trained text embeddings through learned feature mappings~\cite{SPNet, ZS3Net}. 
The introduction of large-scale vision-language models like CLIP~\cite{clip} marked a turning point, empowering subsequent methods with strong zero-shot capabilities. Two major paradigms have since evolved: two-stage and single-stage frameworks.
Two-stage methods~\cite{ovseg, maskclip, odise, han2023global, zegformer, Han2023ZeroShotSS, freeseg} first generate class-agnostic mask proposals, then classify them using text embeddings.
For example, OVSeg~\cite{ovseg} fine-tunes CLIP on masked images paired with text annotations to enhance classification accuracy in the second stage. Similarly, MaskCLIP~\cite{maskclip} incorporates mask tokens into the pre-trained CLIP model to refine masks and improve classification. 
While ODISE~\cite{odise} uses a diffusion model for mask generation and CLIP-based features for recognition.
In contrast, single-stage methods~\cite{lseg, maskclip1, fcclip, catseg, jiao2024collaborative, groupvit, mukhoti2023open} directly align pixel-level features with CLIP's text embeddings. LSeg~\cite{lseg} pioneers this direct alignment, while SAN~\cite{san} and FC-CLIP~\cite{fcclip} enhance it with architectural refinements. Recent works like CAT-Seg~\cite{catseg} and SED~\cite{xie2024sed} improve performance by aggregating cost maps spatially or across scales. Notably, CAT-Seg~\cite{catseg}  further finetunes the CLIP model during training to better adapt it to the segmentation task.
Further, MAFT~\cite{maft} and MAFTPlus~\cite{jiao2024collaborative} explore text encoder fine-tuning strategies, introducing content-aware mechanisms to retain CLIP's generalization while improving segmentation of novel categories.

\subsection{Vision-Language Models}
Visual-language models (VLMs) integrate visual and linguistic modalities to enable open-world visual understanding. Early VLMs~\cite{Li2019VisualBERTAS, Lu2019ViLBERTPT}, trained on small multi-modal datasets, use task combinations for cross-modal alignment. The introduction of CLIP~\cite{clip} marks a shift, utilizing large-scale image-text pairs and contrastive learning for direct alignment, while ALIGN~\cite{align} further scales datasets, embracing noisy data for diverse pretraining. Concurrently, large language models (LLMs) such as LLaMA~\cite{llama} and QWen~\cite{Qwen2.5} have achieved remarkable success in language tasks, providing strong and adaptable foundations. Building on this progress, large multimodal models (LMMs) like LLaVA~\cite{llava} and QWen-VL~\cite{Qwen2.5-VL} have emerged, integrating visual encoders, pre-trained LLMs, and cross-modal alignment modules. Together, these developments have significantly advanced the open-world understanding capabilities of multimodal systems.


\subsection{Step-by-Step Reasoning}
To emulate human sequential reasoning, step-by-step techniques guide Large Language Models (LLMs) in producing structured reasoning steps. The Chain-of-Thought (CoT)~\cite{cot} method enhances arithmetic reasoning but relies on manually crafted prompts. To overcome this limitation, Auto-CoT~\cite{auto-cot} automates demonstration creation with questions and reasoning chains. Building on this, Tree-of-Thoughts (ToT)~\cite{tot} enables LLMs to explore multiple reasoning pathways for flexible decision-making. Further advancing the field, Graph-of-Thoughts (GoT)~\cite{got} models information as a graph, with units as vertices and relationships as edges. Despite these advances, all methods maintain a sequential decoding approach, directing LLMs to reason step-by-step.

Although recent open-vocabulary segmentation methods have achieved remarkable progress, they still struggle to distinguish objects with similar appearance or semantic meaning.
We propose Step-by-Step Visual Reasoning for Open-Vocabulary Segmentation, a framework that guides the segmentation process through interpretable, hierarchical visual explanations via LMMs.

%% file: sec/3_method.tex
\section{Method}

\subsection{Preliminary: Open-Vocabulary Segmentor}
An open-vocabulary segmentor $\mathbf{A}$ is designed to perform segmentation on images that may contain arbitrary, potentially unseen object categories, despite being trained on a fixed set of classes.
Given an input image $\mathbf{o} \in \mathbb{R}^{H \times W \times 3}$ and a target class $c \in \mathcal{C}$, where $\mathcal{C}$ is an open-ended or user-defined set of class names, the goal of the segmentor is to produce a segmentation mask $\mathbf{m}_c$ that highlights regions in the image corresponding to class $c$.
To achieve this, the class name $c$ is first converted into a textual prompt $p_c$ (e.g., \texttt{`a photo of a \{$c$\}'}) using a prompt composer ${P}$.
This prompt is then embedded into a joint visual-semantic space using a vision-language model such as CLIP~\cite{clip}.
Subsequently, the segmentor ${S}$ integrates the text embedding with image features to generate the class-specific mask prediction.
Formally, this process can be formulated as: 
\begin{equation} \mathbf{m}_c = {S}(\mathbf{o}, \textrm{CLIP}(\mathcal{P}_c)), \quad 	\mathcal{P}_c  = {P}(c), \quad c \in \mathcal{C}, \end{equation}
where $\mathbf{m}_c$ denotes the predicted binary mask corresponding to class $c$.

This design allows the segmentor to generalize beyond training classes by conditioning on arbitrary class names.
However, it still suffers struggles in complex scenarios, such as distinguishing between visually or semantically similar objects.
To address these limitations, we propose a Step-by-Step Visual Reasoning framework for Open-Vocabulary Segmentation (\textbf{\ourmodel}).

\begin{figure}[t!]
	\centering
	\includegraphics[width=\textwidth]{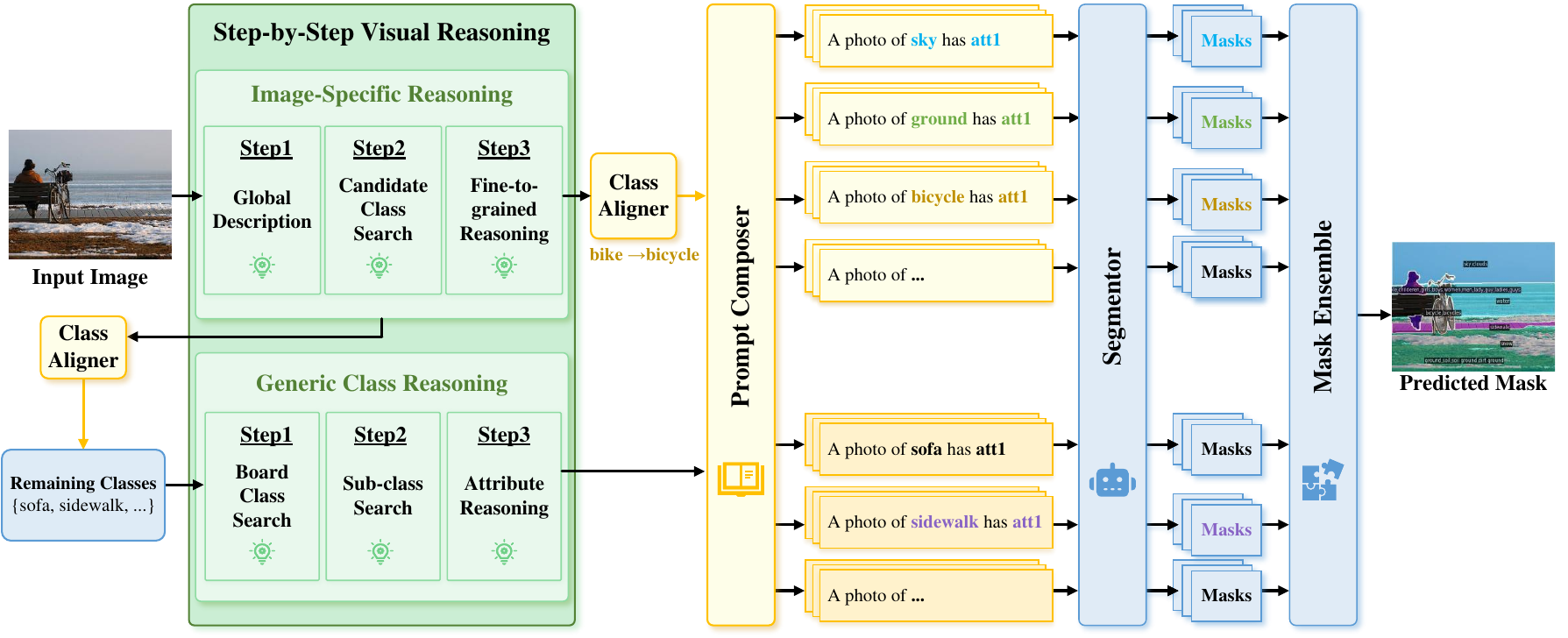}
	\caption{The whole framework of our \textbf{\ourmodel}.
		We first use an LMM to generate image-specific reasoning in a step-by-step manner.
		Predicted classes are aligned to the predefined label set via a class aligner.
		For unobserved classes, the LMM generates generic coarse-to-fine reasoning.
		All reasoning results are composed into prompts and fed to the segmentor to produce masks, which are then aggregated into the final prediction.
	} 
	\label{fig:main}
\end{figure}

\subsection{Step-by-Step Visual Reasoning}
As shown in~\cref{fig:main}, our \textbf{OpenSeg-R} first performs step-by-step visual reasoning to generate image-specific reasoning.
Next, we apply a \textit{Class Aligner} to map the class names predicted by the LMM back to the original class set.
To complement missing categories, we further introduce \textit{Generic Class Reasoning}.
The combined reasoning outputs are then passed to the \textit{Prompt Composer} and \textit{Segmentor}, followed by mask ensembling to produce the final segmentation results. We provide a detailed illustration of each component in the following sections.

\subsubsection{Image-Specific Reasoning}
Instead of directly feeding the image and candidate class set into the segmentor, we first prompt a large multimodal model (LMM), such as Qwen2.5-VL~\cite{Qwen2.5-VL}, to perform a step-by-step analysis of the visual content.
Concretely, the reasoning process consists of three steps.
\textbf{Step 1}: the LMM generates a high-level description of the input image, capturing its overall context and salient content.
This step ensures that the model builds a global understanding of the scene before performing object-level reasoning.
\textbf{Step 2}: conditioned on this description and the image, the LMM identifies which classes from the class set $\mathcal{C}$ are present in the scene, yielding a filtered subset $\mathcal{C}_o$.
This step selects candidate classes, providing the foundation for generating fine-grained class-specific reasoning in the next stage.
\textbf{Step 3}, for each identified class in $\mathcal{C}_o$, the LMM generates a coarse-to-fine reason that justifies its presence in the image.
This process can be formalized as: 
\begin{equation}
\mathcal{T} _o= (\mathbf{o}, \mathcal{C}_o, \mathcal{R}_o), \quad
\text{where} \ 
\begin{cases}
	\mathcal{D} = \text{LMM}(\mathbf{o}), \\
	\mathcal{C}_o = \text{LMM}(\mathbf{o}, \mathcal{D}, \mathcal{C}), \\
	\mathcal{R}_o = \text{LMM}(\mathbf{o}, \mathcal{C}_o).
\end{cases}
\end{equation}
$\mathcal{T} _o$ is the image-specific reasoning triplet, $\mathcal{D}$ denotes the overall image description, $\mathcal{C}_o$ is the subset of classes observed in the image, and $\mathcal{R}_o$ contains the corresponding reasoning outputs for each class in $\mathcal{C}_o$.
For each image, the reasoning result is organized as a triplet $(\mathbf{o}, \mathcal{C}_o, \mathcal{R}_o)$.

Notably, our \ourmodel{} benefits from the step-by-step strategy, as it decomposes the task into manageable sub-problems, making it easier for the LMM to understand and reason effectively.
Furthermore, the coarse-to-fine reasoning process begins with broad categories (e.g., living being), narrows down to subcategories (e.g., animal), and finally identifies the most discriminative semantic attributes (e.g., four legs).
This progressive structure helps the model highlight distinguishing features, enabling it to better differentiate the target class from other visually or semantically similar categories.

\subsubsection{Class Aligner}
Due to the limited context retention and instruction-following consistency of LMMs, the predicted class names may not always exactly match the original input class set.
To address this issue and provide more reliable guidance to the segmentor, we introduce a Class Aligner module that maps the class names generated by the LMM back to the original predefined class set.
Concretely, for class names $c_u$ that are not defined in the original class set $\mathcal{C}$, we employ Sentence Transformers~\cite{SentenceTransformers} (denoted as {SenTF}) to perform semantic alignment.
Specifically, we first encode the LMM-predicted class $c_u$ and all candidate classes $c \in \mathcal{C}$ using SenTF, and compute their pairwise cosine similarity.
We then select the most semantically similar class $c_r$ from the original class set as the matched class: 
\begin{equation}
	c_r = \argmax_{c \in \mathcal{C}} \cos\left[ \text{{SenTF}}(c_u), \text{{SenTF}}(c) \right], c_u \in \mathcal{C}_o, c_u \notin \mathcal{C},
\end{equation}
where $\cos[,]$ denotes the cosine similarity between the {SenTF} embeddings.
We further introduce a similarity threshold $\sigma$  to ensure reliable replacement.
If the maximum similarity score exceeds $\sigma$, we treat $c_r$ as the aligned class for $c_u$; otherwise, $c_u$ is considered unmatched and discarded from subsequent segmentation: 
\begin{equation}
	\hat{c}_u = 
	\begin{cases}
		c_r, & \text{if } \cos\left[ \text{{SenTF}}(c_u), \text{{SenTF}}(c_r) \right] > \sigma; \\
		\emptyset, & \text{otherwise}.
	\end{cases}
\end{equation}

After this alignment step, we obtain the semantic aligned image-specific reasoning triplet: 
\begin{equation}
\widehat{\mathcal{T}}_o = (\mathbf{o}, \widehat{\mathcal{C}}_o, \widehat{\mathcal{R}}_o), 
\end{equation}
where $\widehat{\mathcal{C}}_o$ is the filtered set of matched classes and $\widehat{\mathcal{R}}_o$ contains the corresponding reasoning outputs.
This alignment process ensures consistency between the LMM's generated outputs and the segmentor’s required inputs, thereby improving both the accuracy and robustness of downstream open-vocabulary segmentation.

Although the LMM can recognize most salient objects, it may overlook classes that occupy small regions or have low visual prominence —such as the `sidewalk' in~\cref{fig:main}.
In addition, when the candidate class list is too long, the LMM may ignore certain valid classes due to its limited context length and attention capacity.
Such omissions can result in incomplete or biased segmentation outcomes.
To address this issue, we introduce a Generic Class Reasoning module as a complementary component to the image-specific reasoning.

\subsubsection{Generic Class Reason}
For the classes $c_g\notin \widehat{\mathcal{C}}_o$, that was not detected by the LMM, we further prompt the LMM to provide a generic explanation describing how this class is visually distinguishable from others.
This reasoning is also structured in a coarse-to-fine manner, starting from broad category distinctions and progressively focusing on discriminative fine-grained visual attributes.
After that, we obtain the {generic class reasoning triplet}:
\begin{equation}
\mathcal{T}_g =	\left(\mathbf{o}, \mathcal{C}_g, \mathcal{R}_g\right),
\end{equation}
where \( \mathcal{C}_g \) denotes the set of candidate classes that were not detected in the image-specific reasoning (i.e., \( \mathcal{C}_g = \mathcal{C} \setminus \widehat{\mathcal{C}}_o \)), and \( \mathcal{R}_g \) contains the corresponding generic reasoning outputs generated by the LMM for each class in \( \mathcal{C}_g \).

Overall, the final reasoning instruction is constructed by integrating both the {image-specific} and the {generic class Reasoning} components:
\begin{equation}
	\mathcal{T} = \left(\mathbf{o}, \mathcal{C}, \mathcal{R} \right) 
	= ( \mathbf{o},\ \widehat{\mathcal{C}}_o \cup \mathcal{C}_g,\ \widehat{\mathcal{R}}_o \cup \mathcal{R}_g ).
\end{equation}
With visual reasoning triplet $\mathcal{T}$, which encodes rich semantic and structural information, we then feed it into the segmentor to generate the corresponding segmentation masks.

\subsection{Mask Generation}

\paragraph{Prompt Composer}
We employ a reason prompt composer ${P}_r$ to convert each class-reason pair from $\mathcal{T}$ into a descriptive textual prompt for the segmentor.  
Unlike traditional prompts that consist only of class names (e.g., \texttt{`a photo of \{$c$\}'}), our ${P}_r$ incorporates the associated reasoning to enrich semantic guidance.  
Specifically, for each pair $(c, r)$, we use a compositional template such as \texttt{`a photo of \{$c$\} has \{$r$\}'} to construct the final prompt.
Formally, the composed prompts are generated as:
\begin{equation}  
	\mathcal{P}_c = {P}_r(c, \mathcal{R}_c), \quad c\in \mathcal{C},
\end{equation}
where $\mathcal{P}_c$ denotes the prompt set  for class $c$ using its corresponding reasoning $\mathcal{R}_c$.

\paragraph{Reason-Guided Mask}  
Following standard segmentor~\cite{jiao2024collaborative,xie2024sed}, we use CLIP~\cite{clip} to embed the reason-based prompts as text embeddings. These embeddings are then fed into the segmentor ${S}$ alongside the input image to generate reason-guided segmentation masks:
\begin{equation} 
	\mathbf{m}_c^{r} = {S}(\mathbf{o}, \textrm{CLIP}(\mathcal{P}_c)), \quad \mathbf{m}_c^{r} \in \mathbb{R}^{N \times H \times W },
\end{equation}
where $\mathbf{m}_c^{r}$ denotes the set of predicted masks for class $c$, guided by the corresponding reason-based prompts $\mathcal{P}_c$, and $N$ is the number of  reasons associated with class $c$.

\paragraph{Mask Ensemble}  
To obtain a single unified binary mask for each class, we aggregate the multiple reason-guided masks using average pooling, followed by a sigmoid activation and thresholding.  
For each class \( c \), the final binary segmentation mask is computed as:
\begin{equation}
	\bar{\mathbf{m}}_c = \mathbb{I} \left[ \sigma \left( \frac{1}{N} \sum_{i=1}^{N} \mathbf{m}_c^{r_i} \right) > \tau \right],
\end{equation}
where $\mathbf{m}_c^{r_i} \in \mathbb{R}^{H \times W} $ is the predicted mask from the \( i \)-th reasoning prompt,
$\sigma(\cdot)$ denotes the sigmoid function,
$ \tau $ is a predefined binarization threshold,
$\mathbb{I}[\cdot]$ is the indicator function,
and $\bar{\mathbf{m}}_c \in \mathbb{R}^{H \times W}$ is the final binary segmentation mask, obtained by aggregating the reasoning knowledge associated with the class.

The final predicted mask is inferred based on the visual reasoning generated by the LLM.
This process ensures that segmentation is guided by interpretable, fine-grained visual cues, leading to more precise and semantically aligned predictions.


%% file: sec/4_experiments.tex
\section{Experiments}\label{experiment}
\subsection{Datasets and Evaluation Metric}

\paragraph{Datasets}
Following prior works~\cite{catseg, san, opsnet, odise, jiao2024collaborative}, we train \ourmodel{} on COCO-Stuff~\cite{caesar2018coco} and COCO-Panoptic~\cite{Lin2014MicrosoftCC}. COCO-Stuff provides 164,000 images with semantic annotations across 171 categories, while COCO-Panoptic offers panoptic annotations over the same image set for 133 categories. We use the full training splits of both datasets. To evaluate \ourmodel{}, we test it across several widely-used benchmarks, including ADE20K~\cite{ade20k} (A-150 and A-847), PASCAL VOC~\cite{everingham2010pascal} (PAS-20), and PASCAL-Context~\cite{pascal_context} (PC-59 and PC-459), covering a wide spectrum of categories and scene complexities. ADE20K includes 20,000 training images and supports both 150-category and 847-category open-vocabulary splits. PASCAL-Context extends PASCAL VOC with 4,998 training and 5,005 validation images, and defines PC-59 and PC-459 as open-vocabulary test sets. PASCAL VOC itself contains 20 object categories with augmented annotations for segmentation. Beyond semantic segmentation, we further validate \ourmodel{} in the open-vocabulary panoptic segmentation setting~\cite{opsnet, odise}, evaluating on ADE20K to demonstrate its versatility across tasks.

\paragraph{Evaluation Metric}
Following standard practices in traditional and open-vocabulary semantic segmentation~\cite{zegformer, ovseg, opsnet, odise}, we use mean Intersection over Union (mIoU), calculated as the average intersection over union across all classes, to quantitatively evaluate semantic segmentation performance. For panoptic segmentation, we adopt Panoptic Quality (PQ), Segmentation Quality (SQ), and Recognition Quality (RQ) as evaluation metrics.


\subsection{Implementation Details}
We use \texttt{Qwen2.5-VL-72B-Instruct-AWQ} \cite{Qwen2.5-VL} to process input images and generate step-by-step reasoning outputs, running on 2 A100 GPUs.
The detailed inference prompts are provided in the supplementary material.
In the {Prompt Composer}, we use descriptive attributes as reasoning outputs and combine them with class names to construct prompts for mask generation.
For class name alignment, we employ {all-MiniLM-L6-v2} from Sentence Transformers\cite{SentenceTransformers}, using a similarity threshold of $\sigma=0.5$.
$ \tau $  is also set to 0.5.
Since the PAS-20 dataset contains relatively few candidate classes, the LMM can reliably identify all relevant categories; thus, we omit generic reasoning for this dataset.
We adopt both {SED} \cite{xie2024sed} and {MAFT+} \cite{jiao2024collaborative} as our segmentors, each evaluated using 4 A100 GPUs.
For open-vocabulary semantic segmentation, both models are pretrained on COCO-Stuff~\cite{caesar2018coco}.
For panoptic segmentation, \texttt{MAFT+} is pretrained on COCO-Panoptic~\cite{Lin2014MicrosoftCC}.

\input{tables/0_sota_semantic}

\subsection{Comparison with State-of-the-arts}

\paragraph{Open-Vocabulary Semantic Segmentation}
We compare our method with state-of-the-art approaches on open-vocabulary \textit{semantic} segmentation, as shown in~\cref{tab:main_table}.  
Among them, SED~\cite{xie2024sed} and MAFT+~\cite{jiao2024collaborative} are representative one-stage and two-stage segmentation frameworks, respectively.  
We integrate our reasoning module into both architectures, resulting in two variants: \textbf{\ourmodel} w/ SED and \textbf{\ourmodel} w/ MAFT+.
With a relatively small vision-language model (VLM), our method consistently outperforms the corresponding baselines across all benchmarks.
The proposed \textbf{Open-\ourmodel} variants achieve the best results in all settings.
Specifically, on \texttt{A-150}, our method yields an improvement of approximately \textbf{2\%} when integrated with SED~\cite{xie2024sed}.
On \texttt{A-847}, which contains more fine-grained categories than \texttt{A-150}, \textbf{\ourmodel} w/ MAFT+ achieves a \textbf{1.7\%} gain over MAFT+~\cite{jiao2024collaborative} , surpassing or matching the performance of methods that utilize significantly larger VLMs.
A similar trend is observed on \texttt{PAS-20}, further demonstrating the effectiveness and generalization ability of our reasoning-enhanced framework.
Our method surpasses all SOTA methods on \texttt{A-150}, \texttt{A-847}, and \texttt{PC-459}.

When using larger VLM, our method achieves further improvements across all datasets.
However, the relative gains are smaller compared to that with the smaller VLM.
On \texttt{PC-459} with SED~\cite{xie2024sed} and \texttt{PAS-20} with MAFT+~\cite{jiao2024collaborative}, we even observe a slight performance drop.
This may be attributed to that larger VLMs tend to produce category-level and generic text embeddings, which can suppress attribute-level cues.
Similar issues have been observed in prior work~\cite{fineclip,xie2025fgclip}.
We also find that CAT-Seg~\cite{catseg} is not well-suited for our \ourmodel{}, as its finetuning of CLIP disrupts the visual-semantic space, making it less compatible with our detailed visual reasoning.

\input{tables/0_sota_panoptic}

\paragraph{Open-Vocabulary Panoptic Segmentation}
In~\cref{tab:ovs-pan}, we compare our method on the open-vocabulary \textit{panoptic} segmentation task.
Once again, our approach achieves consistent improvements across all evaluation metrics and outperforms all baselines.
Specifically, it brings a gain of \textbf{0.9\%} in Panoptic Quality (PQ), \textbf{2.3\%} in Segmentation Quality (SQ), and \textbf{1.3\%} in Recognition Quality (RQ).
These results demonstrate the effectiveness of our reasoning-guided framework in enhancing both semantic understanding and instance-level discrimination in open-vocabulary scenarios.

\input{tables/1_ablation1}

\subsection{Ablation Study}
\paragraph{Prompt Type}
The reasoning output consists of three components: broad class, sub-class, and fine-grained visual attributes.
Based on these, we construct four types of prompts: (1) \textit{class name only} (baseline), (2) \textit{coarse reason} (broad + sub-class), (3) \textit{coarse + attribute}, and (4) \textit{attribute only}.
As shown in~\cref{tab:ablation1}, using coarse reasoning alone slightly degrades performance, as such descriptions (e.g., `a dog that is an animal') are too generic to distinguish similar categories.
In contrast, incorporating fine-grained attributes consistently improves results across all datasets.
Notably, prompts using attributes alone achieve the best performance in most cases, suggesting that detailed visual cues provide the most discriminative and informative guidance for open-vocabulary segmentation.

\paragraph{Reason Type}
We evaluate the effectiveness of different reasoning types in~\cref{tab:ablation2}, comparing three configurations: (1) generic class reasoning only, (2) image-specific reasoning only, and (3) their combination.
Generic reasoning performs worse across most datasets, as its descriptions are often too broad to distinguish visually similar objects, leading to lower segmentation accuracy.
In contrast, image-specific reasoning, which is inferred from image content, provides more discriminative cues and achieves better performance.
However, LMMs may still overlook certain objects.
By combining both reasoning types, our method compensates for such omissions: generic reasoning recovers missed categories, while image-specific reasoning improves class-level discrimination.
As shown in the table, this hybrid approach yields the best overall results.
For PAS-20, which contains relatively few and easily recognizable categories, image-specific reasoning alone is sufficient, as the LMM can reliably detect all relevant classes without generic supplementation.
\input{tables/1_ablation2}

\subsection{Qualitative Results}
\begin{figure}[t!]
	\centering
	\includegraphics[width=\textwidth]{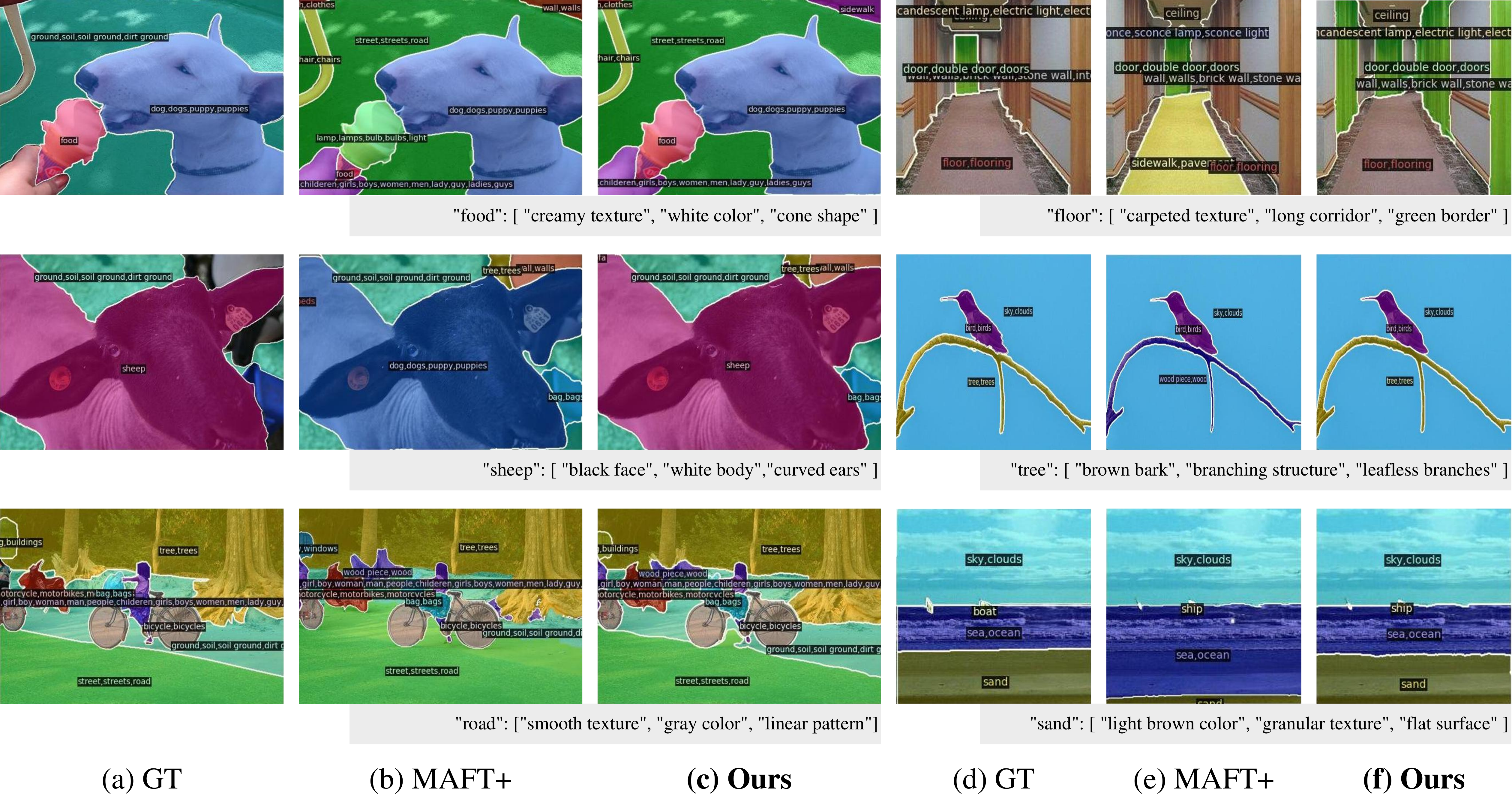}
       \vspace{-3pt}
	\caption{Qualitative comparison with MAFT+~\cite{jiao2024collaborative} on open-vocabulary semantic segmentation.
		We highlight the visual reasoning behind the class that is missed by MAFT, demonstrating how our method recovers it through step-by-step visual reasoning.}
	\label{fig:seg_vis}
       \vspace{-10pt}
\end{figure}

To further illustrate the effectiveness of our method, we present qualitative comparisons in~\cref{fig:seg_vis}.  
We present several  cases where baseline methods MAFT+~\cite{jiao2024collaborative} fail to correctly segment or classify objects due to ambiguous visual or semantic cues.
For each case, we show the visual reasoning attributes inferred by the LMM.
These reasoning cues help disambiguate visually similar categories and guide the segmentor toward more accurate predictions.  
As shown, our method produces more semantically aligned masks across a variety of categories, such as \texttt{floor}, \texttt{tree}, \texttt{road}, and \texttt{sand}, demonstrating the benefit of our \ourmodel{} in open-vocabulary segmentation.

%% file: tables/0_sota_semantic.tex
\begin{table*}[t!]
    \begin{center}
   \caption{Comparison with state-of-the-art methods on open-vocabulary \textit{semantic} segmentation. “\textbf{\ourmodel} w/” indicates our method combined with the corresponding segmentor. The results with improvements are marked in bold.}
    \resizebox{0.75\textwidth}{!}{
    \begin{tabular}{l|c|ccccc}
\toprule
        Method & VLM &  \texttt{A-847} & \texttt{PC-459} & \texttt{A-150} & \texttt{PC-59} & \texttt{PAS-20} 
        \\
\midrule
        SPNet~\cite{SPNet} & - & - & - & - & 24.3 & 18.3  \\
        ZS3Net~\cite{ZS3Net} & - & - & - & - & 19.4 & 38.3 \\
        LSeg~\cite{lseg} & ViT-B/32 &  - & - & - & - & 47.4 \\
        LSeg+~\cite{openseg} & ALIGN & 2.5 & 5.2 & 13.0 & 36.0 & -  \\
        Han et al.~\cite{han2023global} & ViT-B/16 & 3.5 & 7.1 & 18.8 & 45.2 & 83.2  \\
        GroupViT~\cite{groupvit} & ViT-S/16 & 4.3 & 4.9 & 10.6 & 25.9 & 50.7   \\
        ZegFormer~\cite{zegformer} & ViT-B/16 & 4.9 & 9.1 & 16.9 & 42.8 & 86.2 \\
        SimBaseline~\cite{xu2022simple} & ViT-B/16 & 7.0 & - & 20.5 & 47.7 & 88.4\\
        OpenSeg~\cite{openseg} & ALIGN & 4.4 & 7.9 & 17.5 & 40.1 & - \\
        DeOP~\cite{Han2023ZeroShotSS} & ViT-B/16 & 7.1 & 9.4 & 22.9 & 48.8 & 91.7  \\
        PACL~\cite{mukhoti2023open} & ViT-B/16  & - & - & {31.4} & 50.1 & 72.3  \\
        OVSeg~\cite{ovseg} & ViT-B/16 & 7.1 & 11.0 & 24.8 & 53.3 & 92.6  \\
        SAN~\cite{san} & ViT-B/16 & 10.1 & 12.6 & 27.5 & 53.8 & {94.0}   \\
        CAT-Seg~\cite{catseg} & ViT-B/16 & 12.0&19.0&31.8&57.5&94.6 \\
        SCAN~\cite{liu2024open}& ViT-B/16&10.8&13.2&30.8&58.4&97.0\\
        \midrule
        SED~\cite{xie2024sed} & ConvNeXt-B & 11.4  &{18.6} &{31.6}  &{57.3} &  {94.4}\\
    \rowcolor{cyan!10}\textbf{\ourmodel} w  SED& ConvNeXt-B&\textbf{11.8}&\textbf{18.9}&\textbf{33.6}&\textbf{59.0}&\textbf{95.1}\\
                  
          {MAFT+}~\cite{jiao2024collaborative}& ConvNeXt-B& {13.8}&14.6&{34.6}&{57.5}&{95.4}\\
          \rowcolor{cyan!10}\textbf{\ourmodel} w MAFT+& ConvNeXt-B&\textbf{15.2}&\textbf{15.5}&\textbf{35.5}&\textbf{59.0}&\textbf{96.1}\\
          
\midrule
\midrule
        LSeg~\cite{lseg} & ViT-L/14 & - & - & - & - & 52.3  \\
        OpenSeg~\cite{openseg} & ALIGN & 8.1 & 11.5 & 26.4 & 44.8 & - \\
        OVSeg~\cite{ovseg} & ViT-L/14 & 9.0 & 12.4 & 29.6 & 55.7 & 94.5  \\
        Ding \textit{et al.}~\cite{maskclip} & ViT-L/14 & 8.2 & 10.0 & 23.7 & 45.9 & -  \\
        ODISE~\cite{odise} & ViT-L/14 & 11.1 & 14.5 & 29.9 & 57.3 & - \\
        HIPIE~\cite{wang2023hierarchical}&BERT-B~\cite{devlin2018bert}& - & - & 29.0 & 59.3 & -\\
        SAN~\cite{san} & ViT-L/14 & 13.7 & 17.1 & 33.3 & 60.2 & 95.5  \\
        FC-CLIP~\cite{fcclip} & ConvNeXt-L &  14.8 & 18.2 & 34.1 & 58.4 & 95.4  \\
CAT-Seg~\cite{catseg} & ViT-L/14 & 16.0	& 23.8& 37.9& 63.3& 97.0 \\
SCAN~\cite{liu2024open}& ViT-L/14&14.0&16.7 &33.5 &59.3 &97.2\\
\midrule
       SED~\cite{xie2024sed} & ConvNeXt-L & 13.9 & {22.6} & 35.2 & {60.6} & 96.1  \\
    \rowcolor{cyan!10}\textbf{\ourmodel} w  SED& ConvNeXt-L&\textbf{14.3}&{22.0}&\textbf{36.1}&\textbf{61.2}&\textbf{96.3}\\ 
       
        {MAFT+}~\cite{jiao2024collaborative}& ConvNeXt-L&{15.3}&16.7&{36.2}&59.5&{96.4} \\
 \rowcolor{cyan!10}\textbf{\ourmodel} w  MAFT+& ConvNeXt-L&\textbf{16.8}&\textbf{17.1}&\textbf{37.1}&\textbf{60.3}&96.2\\
\bottomrule
    \end{tabular}
    }
    \label{tab:main_table}
    \end{center}
   \vspace{-15pt}
\end{table*}

%% file: tables/0_sota_panoptic.tex
\begin{table}[t]
	\centering
	\caption{Open-vocabulary \textit{panoptic} segmentation performance on ADE20K. PQ, SQ, and RQ are used for evaluation. The best results are highlighted with \textbf{bold}. }
    \vspace{-5pt}
	 \resizebox{0.45\textwidth}{!}{
		\begin{tabular}{l|ccc}
\toprule
			Method& ~~~~PQ~~~~ & ~~~~SQ~~~~ & ~~~~RQ~~~~ \\ 
			\midrule
			FreeSeg \cite{freeseg} & 16.3 & - & - \\
			ODISE \cite{odise} & 22.6 & - & - \\
			MaskCLIP \cite{maskclip} ~~ & 15.1 & 70.4 & 19.2 \\
			OPSNet \cite{opsnet} & 19.0 & 52.4 & 23.0 \\
			FC-CLIP \cite{fcclip} & 21.9 & 71.5 & 26.4 \\
			FC-CLIP* \cite{fcclip} & 26.8 & 71.5 & 32.2 \\
			\midrule
			{MAFT+}~\cite{jiao2024collaborative}& {27.1} & {73.5} & {32.9} \\
			\rowcolor{cyan!10}\textbf{\ourmodel} w  MAFT+&\textbf{28.0}&\textbf{75.0}&\textbf{34.2}\\
			 \bottomrule
		\end{tabular}
		 }
       \vspace{-15pt}
\label{tab:ovs-pan}
\end{table}  

%% file: tables/1_ablation1.tex
\begin{table*}[t!]
    \begin{center}
   \caption{Effect of different prompt choices on open-vocabulary segmentation.}
    \resizebox{0.8\textwidth}{!}{
    \begin{tabular}{l|c|c|ccccc}
    \toprule
        Method & Prompt Type& VLM &  \texttt{A-847} & \texttt{PC-459} & \texttt{A-150} & \texttt{PC-59} & \texttt{PAS-20} 
        \\
        \midrule
{MAFT+}~\cite{jiao2024collaborative}&Class name& ConvNeXt-B& {13.8}&14.6&{34.6}&{57.5}&{95.4}\\
\ourmodel\ w MAFT+&Coarse& ConvNeXt-B&13.4&14.3&34.1&57.2&95.8\\
\ourmodel\  w MAFT+&Coarse + Att& ConvNeXt-B&15.1&15.4&35.3&\textbf{59.0}&95.9\\
\ourmodel\ w MAFT+&Att& ConvNeXt-B&\textbf{15.2}&\textbf{15.5}&\textbf{35.5}&\textbf{59.0}&\textbf{96.1}\\
        \midrule
        {MAFT+}~\cite{jiao2024collaborative}& Class name& ConvNeXt-L&{15.3}&16.7&{36.2}&59.5&\textbf{96.4} \\
\ourmodel\ w  MAFT+&Coarse& ConvNeXt-L&15.1&16.8&35.8&59.2&96.3\\
\ourmodel\ w  MAFT+&Coarse + Att& ConvNeXt-L&16.7&\textbf{17.4}&36.7&60.0&96.3\\
\ourmodel\ w  MAFT+&Att& ConvNeXt-L&\textbf{16.8}&17.1&\textbf{37.1}&\textbf{60.3}&96.2\\
        \bottomrule
    \end{tabular}
    }
    \label{tab:ablation1}
    \end{center}
   \vspace{-15pt}
\end{table*}

%% file: tables/1_ablation2.tex
\begin{table*}[t!]
	\begin{center}
		\caption{Comparison of different reasoning types using MAFT+ as the segmentor.}
		\resizebox{0.65\textwidth}{!}{
			\begin{tabular}{c|c|ccccc}
				\toprule
				Reason& VLM &  \texttt{A-847} & \texttt{PC-459} & \texttt{A-150} & \texttt{PC-59} & \texttt{PAS-20} 
				\\
				\midrule
				Generic class & ConvNeXt-B&14.0&13.6&33.4&57.3&95.8\\
				Image-specific& ConvNeXt-B&13.3&15.4&35.2&58.7&\textbf{96.1}\\
				Both& ConvNeXt-B&\textbf{15.2}&\textbf{15.5}&\textbf{35.5}&\textbf{58.9}&95.1\\
				\midrule
				Generic class & ConvNeXt-L&16.0&15.5&35.5&59.2&95.5\\
				Image-specific& ConvNeXt-L&14.0&15.7&35.9&59.3&\textbf{96.2}\\
				Both& ConvNeXt-L&\textbf{16.8}&\textbf{17.1}&\textbf{37.1}&\textbf{60.3}&95.8\\
				\bottomrule
			\end{tabular}
		}
		\label{tab:ablation2}
	\end{center}
	   \vspace{-10pt}
\end{table*}

%% file: sec/5_conclusion.tex
\section{Conclusion}
In this paper, we introduce a novel method  \textbf{\ourmodel} that improves open-vocabulary segmentation through step-by-step visual reasoning.
\ourmodel leverages Large Multimodal Models (LMMs) to perform hierarchical reasoning based on the input image, generating both image-specific and generic class explanations, which are then composed into descriptive prompts to guide segmentation.
Experiments on open-vocabulary semantic and panoptic segmentation benchmarks demonstrate the effectiveness and broad applicability of our approach.


%% file: sec/7_supp.tex
\section{Qualitative Results of Open-Vocabulary Panoptic Sementation}
\begin{figure}[h]
	\centering
	\includegraphics[width=\textwidth]{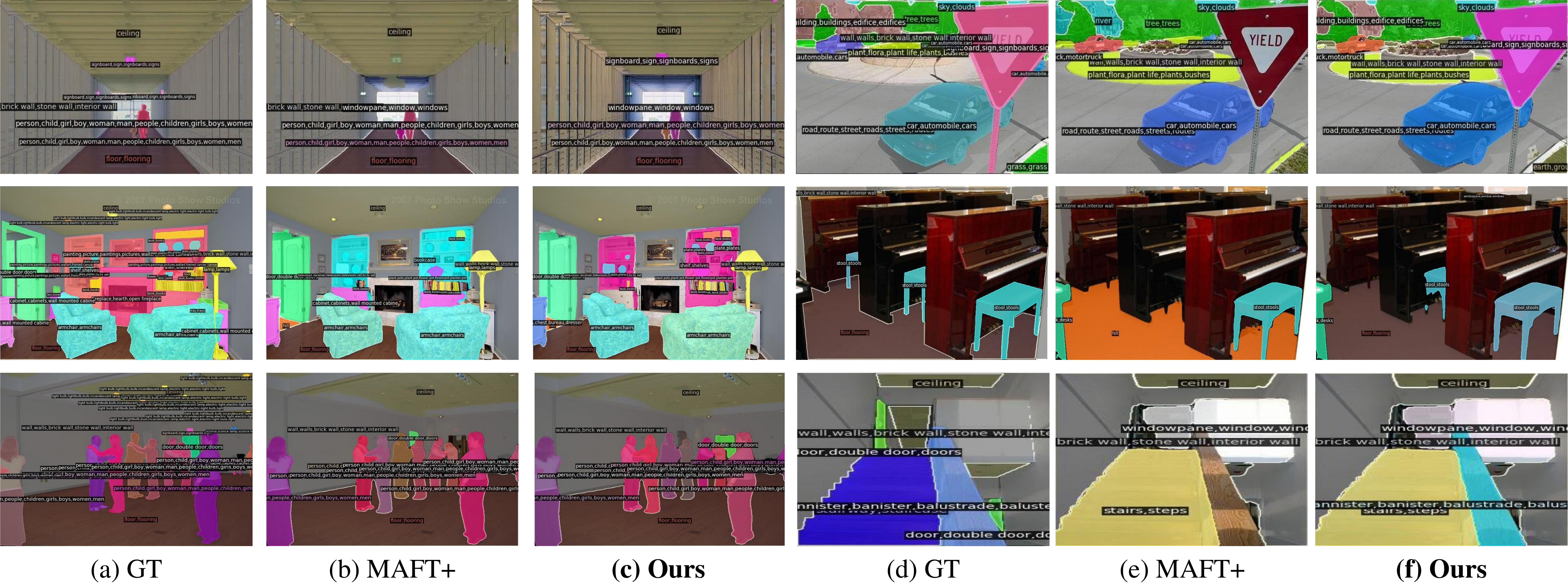}
	\caption{Qualitative comparison with MAFT+~\cite{jiao2024collaborative} on open-vocabulary panoptic segmentation.}
	\label{fig:pano_vis}
\end{figure}

In~\cref{fig:pano_vis}, we present qualitative comparisons with MAFT+~\cite{jiao2024collaborative} on open-vocabulary panoptic segmentation.
Our method consistently delivers superior performance across most cases.
Notably, it demonstrates improved instance-level discrimination. For example, it more accurately identifies ambiguous categories such as \texttt{stool}, and small objects like \texttt{plates}.

\section{More Implementation Details}

\begin{figure}[t]
	\centering
	\includegraphics[width=\textwidth]{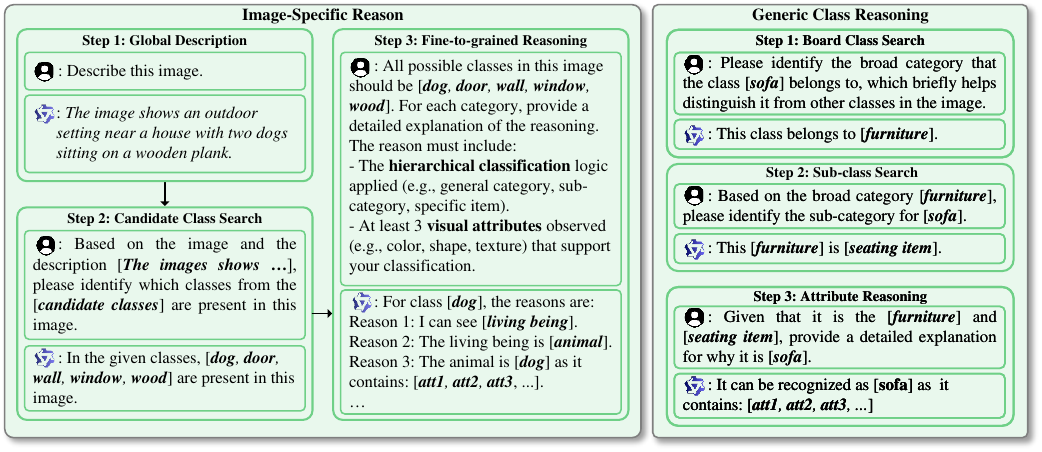}
	\caption{Prompt design for step-by-step visual reasoning. The left panel shows the three-stage process for image-specific reasoning, while the right panel illustrates the prompts of generic class reasoning. }
	\label{fig:reason_step}
\end{figure}

\paragraph{Prompt to LMM} We adopt Qwen2.5-VL-72B-Instruct~\cite{Qwen2.5-VL} as our Large Multimodal Model (LMM) to generate image-specific reasoning.
This model is chosen for its strong instruction-following ability and robust image understanding.
Its responses are generally well-structured and coherent, which makes it easier to perform downstream processing and reasoning alignment.
 We provide the detailed prompts used with Qwen2.5-VL-72B-Instruct~\cite{Qwen2.5-VL} to perform step-by-step visual reasoning in our framework.
As illustrated in \cref{fig:reason_step}, the reasoning process consists of two components: image-specific reasoning and generic class reasoning.
For image-specific reasoning, we guide the LMM through three stages: (1) generating a global image description, (2) identifying candidate classes present in the scene, and (3) producing hierarchical, fine-to-grained reasoning for each observed class.
For generic class reasoning—used to supplement missed classes—we similarly apply a three-step process that involves coarse category identification, sub-category refinement, and attribute-based justification.
This design ensures that the reasoning output is structured, interpretable, and directly usable for composing descriptive prompts in the segmentation stage.
During the use of Qwen2.5-VL-72B-Instruct~\cite{Qwen2.5-VL}, we set the decoding temperature to 0.7.
In a few cases, Qwen2.5-VL-72B-Instruct~\cite{Qwen2.5-VL} fails to recognize certain objects in the image despite multiple attempts.
For these instances, we discard the image-specific reasoning and rely solely on the generic class reasoning to supplement the segmentation process.

\paragraph{Details of Segmentor} 
As noted earlier, we adopt both {SED} \cite{xie2024sed} and {MAFT+} \cite{jiao2024collaborative} as our segmentors.
We follow all implementation settings from their respective official releases.
Specifically, {SED} uses 80 segmentation prompts in the prompt composer to predict masks for each category, while {MAFT+} uses 14 prompts.
The original prompts follow the format `\texttt{a photo of \{$c$\}}', where $c$ denotes the class name. In our method, we enrich these prompts with visual reasoning by appending fine-grained attributes. The modified reason-based format becomes `\texttt{a photo of \{$c$\} that has \{$r$\}}', where $r$ represents the fine-grained attributes generated by the LMM.
We then generate one mask per attribute and apply mask ensembling to produce the final segmentation mask for each class.

